    \documentclass[10pt,twocolumn,letterpaper]{article}
    
    \usepackage[pagenumbers]{cvpr} 
    
    \usepackage{graphicx}
    \usepackage{amsmath}
    \usepackage{amssymb}
    \usepackage{booktabs}
    
    \usepackage{multirow}
    \usepackage{makecell}
    \usepackage{xspace}
    \usepackage{pifont}
    \usepackage[dvipsnames]{xcolor}
    
    \usepackage{threeparttable}

\usepackage{enumitem}

\usepackage{booktabs}
\usepackage{multirow}
\usepackage{makecell}
\usepackage[normalem]{ulem}

\usepackage{color, colortbl}
\definecolor{iGray}{gray}{0.9}
\definecolor{beaublue}{rgb}{0.74, 0.83, 0.9}
\definecolor{Royal_Blue}{rgb}{0.0, 0.1, 0.66}
\usepackage[pagebackref=true,breaklinks=true,letterpaper=true,colorlinks,bookmarks=false, citecolor = Royal_Blue]{hyperref}

\newcommand\blfootnote[1]{%
\begingroup 
\renewcommand\thefootnote{}\footnote{#1}%
\addtocounter{footnote}{-1}%
\endgroup 
}

    \newcommand{\cmark}{\ding{51}}

    \usepackage[capitalize]{cleveref}
    \crefname{section}{Sec.}{Secs.}
    \Crefname{section}{Section}{Sections}
    \Crefname{table}{Table}{Tables}
    \crefname{table}{Tab.}{Tabs.}

    \def \swintiny      {Swin-T\xspace}
    
    \def \swinbase      {Swin-B\xspace}

    \def \deitsmall     {DeiT-S\xspace}
    \def \deitbase      {DeiT-B\xspace}

    \def \deitbasedis   {DeiT-B\alambic\xspace}
    \def \deitbaseup    {DeiT-B$\uparrow$}

    \def \deitbasedisup {DeiT-B\alambic$\uparrow$}

    \def \alambic {\includegraphics[width=0.02\linewidth]{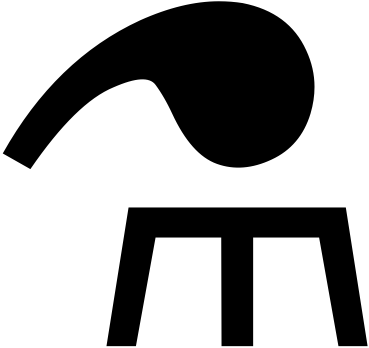}\xspace}

    \def\cvprPaperID{****} 
    \def\confName{CVPR}
    \def\confYear{2022}
    
    \usepackage[accsupp]{axessibility}
    
\begin{document}
    
    \setlength{\belowdisplayskip}{7.2pt} \setlength{\belowdisplayshortskip}{7.2pt}
\setlength{\abovedisplayskip}{7.2pt} \setlength{\abovedisplayshortskip}{7.2pt}

    \title{
    MiniViT: Compressing Vision Transformers with  Weight Multiplexing
    }
 
    \author{Jinnian Zhang$^{1, *}$, Houwen Peng$^{1, *, \dagger}$, Kan Wu$^{1, *}$, Mengchen Liu$^{2}$, Bin Xiao$^{2}$, Jianlong Fu$^{1}$, Lu Yuan$^{2}$\\ 
    $^1$ Microsoft Research,  $^2$ Microsoft Cloud+AI\\
    {\normalsize \{v-jinnizhang, houwen.peng, v-kanwu, mengcliu, bin.xiao, jianf, luyuan\}@microsoft.com}
    }
    \maketitle

    \blfootnote{
  $^*$Equal contributions. Work done when Jinnian and Kan were interns of Microsoft. ~$^\dagger$Corresponding author.
 }
 
\vspace{-5mm}
\begin{abstract}
Vision Transformer (ViT) models have recently drawn much attention in computer vision due to their high model capability. However, ViT models suffer from huge number of parameters, restricting their applicability on devices with limited memory. To alleviate this problem, we propose MiniViT, a new compression framework, which achieves parameter reduction in vision transformers while retaining the same performance. The central idea of MiniViT is to multiplex the weights of consecutive transformer blocks. More specifically, we make the weights shared across layers, while imposing a transformation on the weights to increase diversity. Weight distillation over self-attention is also applied to transfer knowledge from large-scale ViT models to weight-multiplexed compact models. Comprehensive experiments demonstrate the efficacy of MiniViT, showing that it can reduce the size of the pre-trained Swin-B transformer by 48\%, while achieving an increase of 1.0\% in Top-1 accuracy on ImageNet. Moreover, using a single-layer of parameters, MiniViT is able to compress DeiT-B by 9.7 times from 86M to 9M parameters, without seriously compromising the performance. Finally, we verify the transferability of MiniViT by reporting its performance on downstream benchmarks. Code and models are available at
\href{https://github.com/microsoft/Cream}{here}.
    \end{abstract}
    
    \vspace{-3mm}
    \section{Introduction}
    \label{sec:intro}
    
\textit{``Only Mini Can Do It."}

\rightline{\textit{--- BMW Mini Cooper}}

Large-scale pre-trained vision transformers, such as ViT~\cite{ViT}, CvT~\cite{MicrosoftCvT}, and Swin~\cite{Swin}, have recently drawn a great deal of attention due to their high model capabilities and superior performance on downstream tasks. However, they generally involve giant model sizes and large amounts of pre-training data. For example, ViT uses 300 million images to train a huge model with 632 million parameters, achieving state-of-the-art performance on image classification~\cite{ViT}. Meanwhile, the Swin transformer uses 200-300 million parameters, and is pre-trained on ImageNet-22K~\cite{imagenet}, to attain promising results on downstream detection and segmentation tasks~\cite{Swin}.

Hundreds of millions of parameters consume considerable storage and memory, making these models unsuitable for applications involving limited computational resources, such as edge and IoT devices, or in which real-time predictions are needed. Recent studies reveal that the large-scale pre-trained models are over-parametrized~\cite{kovaleva2019revealing} 
. Therefore, it is necessary and feasible to eliminate redundant parameters and the computational overhead of these pre-trained models without compromising their performance.

    \begin{figure}[t]
       \vspace{-0.5cm}
        \hspace{-0.3cm}
        \includegraphics[width=9.4cm]{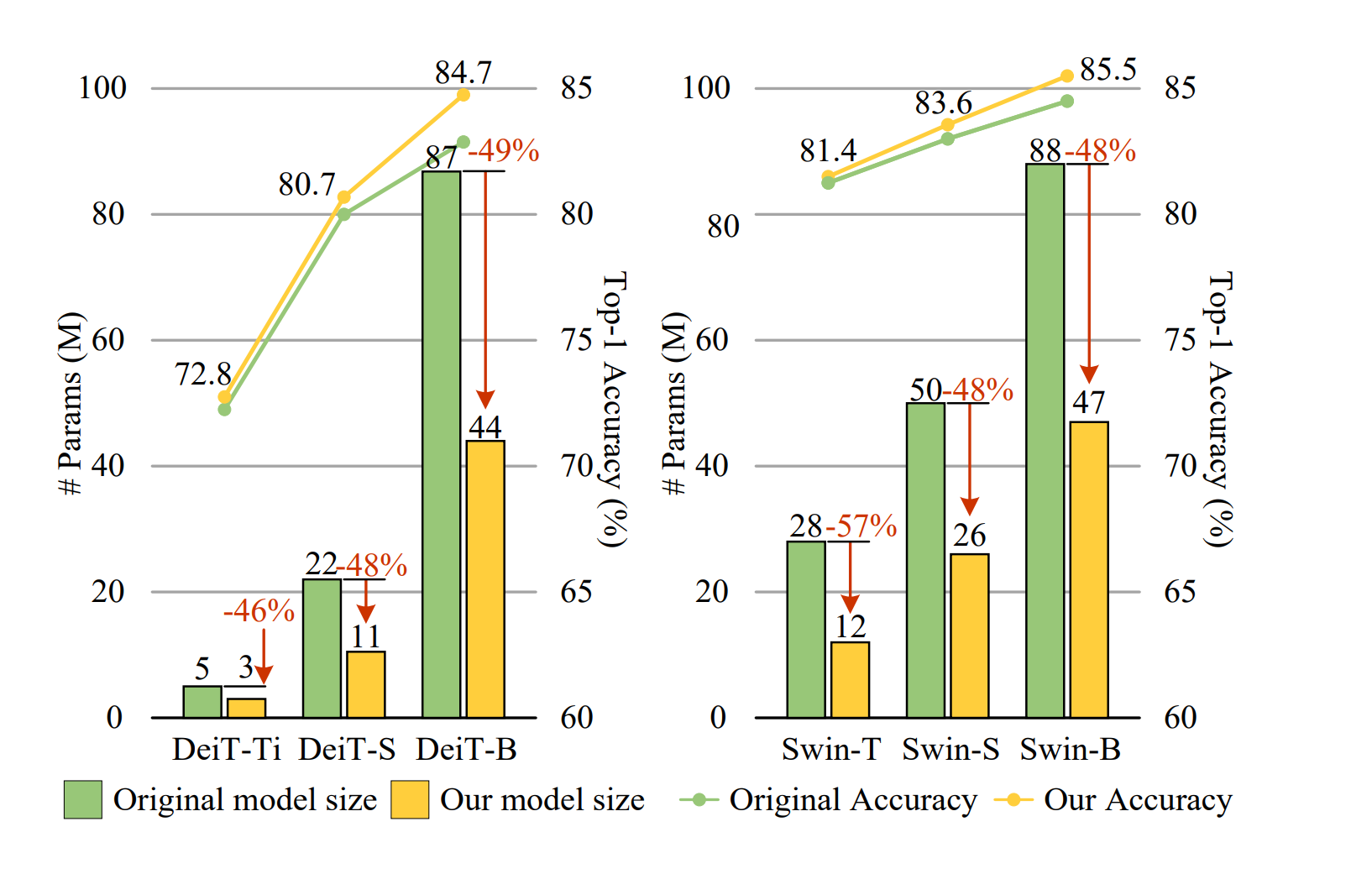}
       \vspace{-1.0cm}
        \caption{Comparisons between MiniViTs and popular vision transformers, such as DeiT~\cite{deit} and Swin Transformers~\cite{Swin}} 
        \vspace{-6mm}
    
        \label{fig:acc_params}
    \end{figure}

Weight sharing is a simple, but effective, technique to reduce model sizes.~The original idea of weight sharing in neural networks was proposed in the 1990s by LeCun and Hinton\cite{ws_lecun, ws_hinton}, and recently reinvented for transformer model compression in natural language processing (NLP) \cite{Albert}.~The most representative work, ALBERT \cite{Albert}, introduces a cross-layer parameter sharing method to prevent the number of parameters from growing with network depth. Such technique can significantly reduce the model size without seriously hurting performance, thus improving parameter efficiency. However, the efficacy of weight sharing in vision transformer compression is not well explored. 

To examine this, we perform the cross-layer weight sharing \cite{Albert} on DeiT-S \cite{deit} and Swin-B \cite{deit} transformers. Unexpectedly, this straightforward usage of weight sharing brings two severe issues: (1) \emph{Training instability.} We observed that weight sharing across transformer layers makes the training become unstable, and even causes training collapse as the number of shared layers increases, as visualized in Fig.~\ref{fig:grad_norm}. (2) \emph{Performance degradation.} The performance of weight-shared vision transformers drops significantly compared to the original models. For example, it leads to a 5.6\% degradation in accuracy for Swin-S, although weight sharing can reduce the number of model parameters by fourfold.

To investigate the underlying reasons for these observations, we analyze the $\ell_2$-norm of gradients during training and the similarities between intermediate feature representations from the model before and after weight sharing (cf. Sec.~\ref{sec: analysis}).~We found that strictly identical weights across different layers is the main cause of the issues. In particular, the layer normalization \cite{LN} in different transformer blocks should not be identical during parameter sharing, because the features of different layers have various scales and statistics.~Meanwhile, the $\ell_2$-norm of the gradient becomes large and fluctuates across different layers after weight sharing, leading to training instability.~Finally, the Central Kernel Alignment (CKA)~\cite{cka} values, a popular similarity metric, drop significantly in the last few layers, indicating that feature maps generated by the model before and after weight sharing become less correlated, which can be the reason of performance degradation.

In this paper, we propose a new technique, called \emph{weight multiplexing}, to address the above issues. It consists of two components, weight transformation and weight distillation, to jointly compress pre-trained vision transformers. 
The key idea of \emph{weight transformation} is to impose transformations on the shared weights, such that different layers have slightly different weights, as shown in Fig.~\ref{fig:diagram}.
This operation can not only promote parameter diversity, but also improve training stability. More concretely, we impose simple linear transformations on the multi-head self-attention (MSA) module and the multilayer perceptron (MLP) module for each weight-shared transformer layer. Each layer includes separate transformation matrices, so the corresponding attention weights and outputs of MLP are different across layers. The layer normalization for different layers is also separated, in contrast to sharing the same parameters. As such, the optimization of weight sharing transformer networks becomes more stable, as demonstrated in Fig.~\ref{fig:grad_norm}.

To mitigate performance degradation, we further equip weight multiplexing with \emph{weight distillation}, such that the information embedded in the pre-trained models can be transferred into the weight-shared small ones, which are much more compact and lightweight. In contrast to previous works that only rely on prediction-level distillation \cite{deit, jia2021efficient}, our method additionally considers both attention-level and hidden-state distillation, allowing the smaller
model to closely mimic the behavior of the original pre-trained large teacher model.

The experiments demonstrate that our weight multiplexing method achieves clear improvements in accuracy over the baselines and compresses pre-trained vision transformers by 2 times while transferring well to downstream tasks. For instance, with the proposed weight multiplexing, the Mini-Swin-B model with 12-layer parameters obtains 0.8\% higher accuracy than the 24-layer Swin-B. Moreover, Mini-DeiT-B with 9M parameters achieves 79.8\% top-1 accuracy on ImageNet, being 9.7 times smaller than DeiT-B (with 86 parameters and 81.8\% accuracy). The 12M tiny model compressed by our approach transfers well to downstream object detection, achieving an AP of 48.6 on the COCO validation set, which is on par with the original Swin-T using 28M parameters.

We summarize our contributions as follows:
\vspace{-2mm}
\begin{itemize}[leftmargin=0.468cm]
	\item{We systematically investigate the efficacy of weight sharing in vision transformers, and analyze the underlying reasons of issues brought by weight sharing.}
\vspace{-2mm}
	\item{We propose a novel compression framework termed MiniViT for general vision transformers. Experimental results demonstrate that MiniViT can achieve a large compression ratio without losing accuracy. Furthermore, the performance of MiniViT transfers well to downstream benchmarks.}

\end{itemize}
\section{Background}
Before presenting our method, we first briefly review some background on vision transformers and parameter sharing, which are fundamental to this work.

\subsection{Vision Transformers}
Transformers, though originally designed for NLP \cite{vaswani2017attention, bert, Albert}, have recently demonstrated their great potentials in computer vision \cite{ViT,deit,Swin}. 
Vision transformers first split an input image into a sequence of 2D patches known as tokens. 
They then flatten and transform these patches to $D$-dimensional vectors using a linear projection \cite{ViT} or stacked CNN layers \cite{T2TViT}, also known as patch embeddings. To retain positional information, positional embeddings are added to patch embeddings.
The combined embeddings are then fed to a transformer encoder (described below). Lastly, a linear layer is used to produce the final classification.

A transformer encoder consists of alternating blocks of \textit{multihead self-attention} (MSA) and \textit{multi-layer perceptron} (MLP) blocks. Layer normalization (LN)~\cite{LN} and residual connections are applied before and after each block, respectively. We elaborate on the MSA and MLP blocks as below.

\noindent
\textit{MSA:}
Let $M$ be the number of heads, also known as self-attention modules. Given the input sequence $\mathbf{Z}_0\in\mathbb{R}^{N\times D}$, in the $k^{th}$ head, we generate queries, keys, and values by linear projections, denoted by $\mathbf{Q}_k$, $\mathbf{K}_k$, and $\mathbf{V}_k\in\mathbb{R}^{N\times d}$ respectively, where $N$ is the number of tokens. $D$ and $d$ are the dimensions of patch embeddings and \textit{Q-K-V} matrices, respectively. We then compute a weighted sum over all values for each position in the sequence. The weights, called attentions and denoted by $\mathbf{A}_k$, are based on the pairwise similarity between two elements in the sequence, namely
\vspace{-3mm}
\begin{align}
    &\mathbf{h}_k=\mathbf{A}_k\mathbf{V}_k, \text{and} \label{eq: value} \\
    &\mathbf{A}_k=\text{softmax}\left(\frac{\mathbf{Q}_k\mathbf{K}_k^T}{\sqrt{d}}\right), \label{eq: attn}
\end{align}    
where \textit{softmax}($\cdot$) is conducted on each row of the input matrix. Finally, a fully-connected layer is applied to the concatenation of the outputs of all heads. 

\noindent
\textit{MLP: }
The MLP block comprises two fully-connected layers with an activation function denoted by $\sigma(\cdot)$, usually GELU \cite{gelu}. Let $\mathbf{Y}\in\mathbb{R}^{N\times d}$ be the input of MLP. The output of MLP can be expressed as
\begin{equation}
    \mathbf{H}=\sigma(\mathbf{Y}\mathbf{W}^{(1)}+\mathbf{b}^{(1)})\mathbf{W}^{(2)}+\mathbf{b}^{(2)}, \label{eq: hidden}
\end{equation}
where $\mathbf{W}^{(1)}\in\mathbb{R}^{d\times d'}$, $\mathbf{b}^{(1)}\in\mathbb{R}^{d'}$, $\mathbf{W}^{(2)}\in\mathbb{R}^{d'\times d}$, and $\mathbf{b}^{(2)}\in\mathbb{R}^{d}$ are the weights and biases for the first and second layers, respectively.
Notably, we usually set $d'>d$.

\begin{figure}[t]
         \vspace{-3mm}
\centering
        \includegraphics[width=8cm]{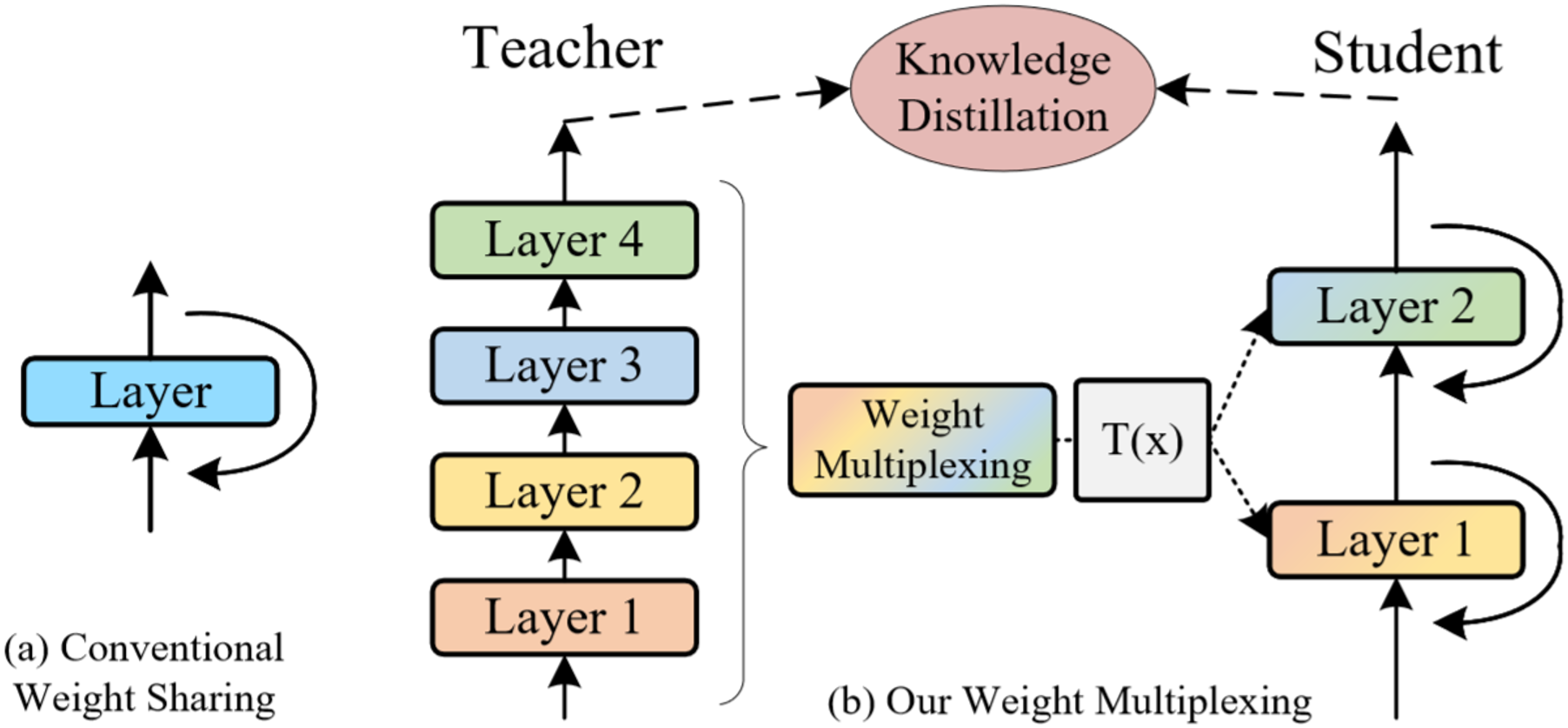}
        \vspace{-0.2cm}
        \caption{Classical weight sharing versus weight multiplexing.} 
        \vspace{-5mm}
    
        \label{fig:diagram}
    \end{figure}

\subsection{Weight Sharing}
    
    Weight sharing is a simple but effective way to improve parameter efficiency. The core idea is to share parameters across layers, as shown in Fig.~\ref{fig:diagram}(a). Mathematically, weight sharing can be formulated as a recursive update of one transformer block  $f$ (i.e., one shared layer):
    \begin{equation}
        \mathbf{Z}_{i+1}=f(\mathbf{Z}_i; \boldsymbol\theta),\quad i=0,\dots, L-1, \label{eq: weight_sharing}
    \end{equation}
    where ${\bf{Z}}_i$ denotes the feature embedding of the sequence in layer $i$, $L$ is the total number of layers, and $\boldsymbol\theta$ represents the shared weights of the transformer block across all layers. The efficacy of weight sharing has been explored and proved in natural language transformer models \cite{DEQ, UniTransformer, Albert}.  It can prevent the number of parameters from growing with the depth of the  network without seriously hurting the performance, thus improving parameter-efficiency.
    
\section{Method}

In this section, we describe our proposed weight multiplexing strategy for vision transformer compression. It consists of two key components, weight transformation and weight distillation, to improve training stability and model performance during weight sharing. Finally, we depict the pipeline of model compression with weight multiplexing.

\subsection{Weight Multiplexing}
\label{sec: wm}

The potential of weight sharing has been demonstrated in NLP \cite{Albert, DEQ, UniTransformer}; however, its efficacy is still unclear in vision transformers. 
To examine this, we directly apply the cross-layer weight sharing in Eq.~(\ref{eq: weight_sharing}) on the DeiT-S \cite{deit} and Swin-B \cite{Swin} transformer models, and observe two issues: training instability and performance degradation. Based on our analysis in Sec.~\ref{sec: analysis}, the strict identity of weights across different layers is the main cause of the issues. In particular, the $\ell_2$-norm of the gradients after weight sharing becomes large and fluctuates in different transformer blocks, as shown in Fig.~\ref{fig:grad_norm}. Furthermore, the CKA values indicate that feature maps generated by the model after weight sharing are less correlated with the original model, as shown in Fig.~\ref{fig:feat_cka}. To solve these issues, inspired by the multiplexing technologies in telecommunications \cite{wiki:multiplexing, cdm}, we propose a new technique called \emph{weight multiplexing} for transformer compression. It combines multi-layer weights into a single weight over a shared part, while involving transformation and distillation to increase parameter diversity. 

More concretely, as shown in Fig.~\ref{fig:diagram}(b),
the proposed weight multiplexing method consists of (1) sharing weights across multiple transformer blocks, which can be considered as a combination process in multiplexing \cite{cdm};  (2) introducing transformations in each layer to mimic demultiplexing \cite{cdm}; and (3) applying knowledge distillation to increase the similarity of feature representations between the models before and after compression. Following Eq.~(\ref{eq: weight_sharing}), we can re-formulate our weight multiplexing as follows:
\begin{equation}
    \mathbf{Z}_{i+1}=f(\mathbf{Z}_i;\boldsymbol\theta,\boldsymbol{\theta}'_i),\ i=0, \dots, L-1,
\end{equation}
where $\boldsymbol{\theta}'_i$ represents the weights of transformation blocks in the $i^{th}$ transformer layer. Note that the number of parameters in $\boldsymbol{\theta}'_i$ is far fewer than $\boldsymbol\theta$.
    
\vspace{-2mm}
\subsubsection{Weight Transformation}
    \begin{figure*}[t]
        \vspace{-1.1cm}
        \hspace{-0.6cm}
        \centering
        \includegraphics[width=16.0cm]{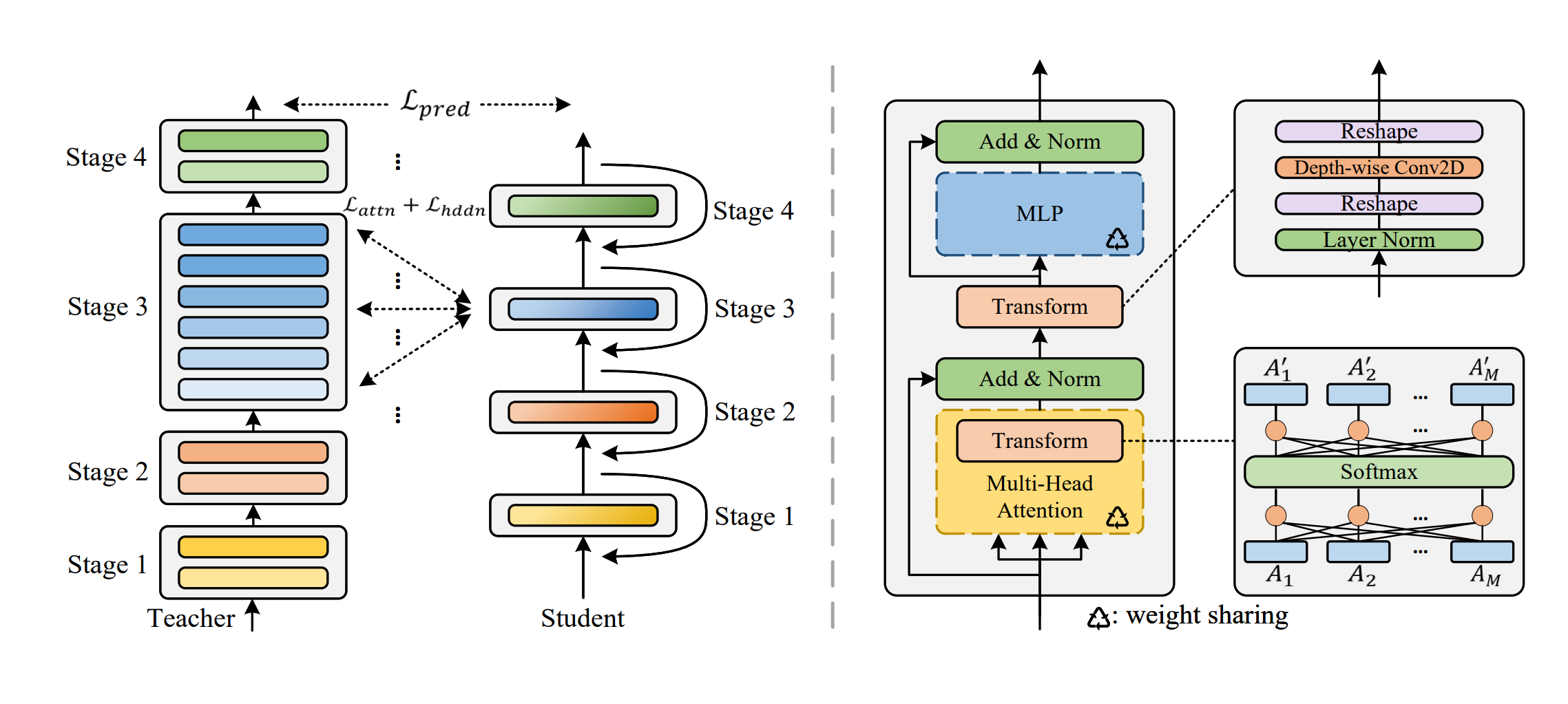}
        \vspace{-1cm}
        \caption {\textbf{Left:} The overall framework of MiniViT. Note that the number of stages are configurable, instead of being fixed in Swin transformers~\cite{Swin}. The transformer layers in each stage of the original models to be compressed should have identical structures and dimension. \textbf{Right:} The detailed transformer block in a MiniViT. We share weights of MSA and MLP in each stage, and add two transformation blocks to increase the parameter diversity. The transformation blocks and normalization layers are not shared. 
        }
        \vspace{-5mm}
    
        \label {fig:transformation}
    \end{figure*}
    
    The transformations in our method are imposed on both attention matrices and feed-forward networks. Such transformations allow each layer to be different, thus elevating parameter diversity and model representation capability. As illustrated in Fig.~\ref{fig:transformation} (Right), the parameters of transformation kernels are \emph{not} shared across layers, while all the other blocks in the original transformer are shared except LayerNorm~\cite{LN}. Since the shared blocks occupy the vast majority of the model parameters,
    the model size only increases slightly after weight multiplexing.

    \textit{Transformation for MSA}. To improve parameter diversity, we insert two linear transformations before and after the softmax self-attention module, respectively. Formally, different from the original self-attention in Eq.~(\ref{eq: value}-\ref{eq: attn}), the transformation-equipped attention is defined as 
    \vspace{-2mm}
    \begin{align}
        & \mathbf{h}_k=\mathbf{A}'_k\mathbf{V}_k=\sum_{n=1}^M\mathbf{F}_{kn}^{(1)}\mathbf{A}_n\mathbf{V}_k  \label{eq:t_mha_before},\\
        &\mathbf{A}_n=\text{softmax}\left(\sum_{m=1}^M\mathbf{F}_{nm}^{(2)}\frac{\mathbf{Q}_m\mathbf{K}_m^T}{\sqrt{d}}\right), \label{eq:t_mha_after}
    \end{align}
    where $\textbf{F}^{(1)}$, $\textbf{F}^{(2)}\in\mathbb{R}^{M\times M}$ are the linear transformation kernels before and after the \emph{softmax}, respectively. Such linear transformations can make each attention matrix $\mathbf{A}_n$ different, while combining information across attention heads to increase parameter variance. 
    
    \textit{Transformation for MLP}. On the other hand, we further impose a lightweight transformation for MLP to elevate parameter diversity. In particular, let the input be $\mathbf{Y}=[\mathbf{y}_1, \dots, \mathbf{y}_d]$, where $\mathbf{y}_l$ denotes the $l^{th}$ position of embedding vectors of all tokens. We then introduce $d$ linear transformations to convert $\mathbf{Y}$ into $\mathbf{Y}'=[\mathbf{C}^{(1)}\mathbf{y}_1, \dots ,\mathbf{C}^{(d)}\mathbf{y}_d]$, where $\mathbf{C}^{(1)}, \dots ,\mathbf{C}^{(d)}\in\mathbb{R}^{N\times N}$ are independent weight matrices of linear layers. Then Eq.~(\ref{eq: hidden}) is re-formulated as
    \vspace{-5mm}
    \begin{align}
        & \mathbf{H}=\sigma(\mathbf{Y}'\mathbf{W}^{(1)}+\mathbf{b}^{(1)})\mathbf{W}^{(2)}+\mathbf{b}^{(2)}. \label{eq: hidden_ours}
    \end{align}
    To reduce the number of parameters and introduce locality in the transformations, we resort to depth-wise convolution~\cite{xception} to sparsify and share weights in each weight matrix, leading to only $K^2d$ parameters compared to $N^2d$ parameters ($K<<N$), where $K$ is the kernel size of convolution. After the transformations, the outputs of MLP become more diverse, improving the parameter efficacy.
    
    Theoretically, with these transformations, the weight-shared layers can restore the behaviors of the pre-trained models, similar to a demultiplexing process. Then the training instability and performance drop issues can be alleviated, because these issues are not observed in the original models. Similar transformations are also applied to improve the performance of transformers without sharing blocks, such as talking-heads attention~\cite{talking_head} and CeiT~\cite{ceit}. However, we extend the ability of transformations to circumvent the drawbacks of weight-sharing methods.
    
\vspace{-0.3cm}
\subsubsection{Weight Distillation}
To compress the large pre-trained models and address the performance degradation issues induced by weight sharing, we further resort to weight distillation to transfer knowledge from the large models to the small and compact models. We consider three types of distillation for transformer blocks, i.e., prediction-logit distillation, self-attention distillation, and hidden-state distillation.

\textit{Prediction-Logit Distillation}. Hinton et al.~\cite{distill_hinton} firstly demonstrated that deep learning models can achieve better performance by imitating the output behavior of well-performing teacher models during training.
We leverage this idea to introduce a prediction loss, as follows:
    \begin{equation}
        \mathcal{L}_{pred} = CE\left(\text{softmax}\left(\frac{\mathbf{z}_s}{T}\right), \text{softmax}\left(\frac{\mathbf{z}_t}{T}\right)\right),
    \end{equation}
where $\mathbf{z}_s$ and $\mathbf{z}_t$ are the logits predicted by the student and teacher models, respectively, and $T$ is a temperature value which controls the smoothness of logits. In our experiments, we set $T=1$. CE denotes the cross-entropy loss.
    
\textit{Self-Attention Distillation}. 
Recent literature has shown that utilizing attention maps in transformer layers to guide the training of student models is beneficial \cite{Albert, tinybert, mobilebert}. To solve the dimension inconsistency between the student and teacher models due to differing numbers of heads, and inspired by \cite{minilmv2}, we apply cross-entropy losses on relations among queries, keys, and values in MSA. 

In particular, we first append matrices over all heads. For example, we define $\mathbf{Q}=[\mathbf{Q}_1, \dots ,\mathbf{Q}_M]\in\mathbb{R}^{N\times Md}$, and $\mathbf{K}$, $\mathbf{V}\in\mathbb{R}^{N\times Md}$ in the same way. For notational simplicity, we write $\mathbf{S}_{1}$, $\mathbf{S}_{2}$, and $\mathbf{S}_{3}$ to denote $\mathbf{Q}$, $\mathbf{K}$, and $\mathbf{V}$ respectively. Then we can generate nine different relation matrices defined by $\mathbf{R}_{ij}=\text{softmax}(\mathbf{S}_{i}\mathbf{S}_{j}^T/\sqrt{Md})$. Note that $\mathbf{R}_{12}$ is the attention matrix $\mathbf{A}$. The self-attention distillation loss can be expressed as
    \vspace{-2mm}
    \begin{equation}
        \mathcal{L}_{attn}=\frac{1}{9N}\sum_{n=1}^N\sum_{i,j\in\atop\{1,2,3\}}CE(\mathbf{R}_{ij,n}^{s}, \mathbf{R}_{ij,n}^{t}),
    \end{equation}
    where $\mathbf{R}_{ij,n}$ represents the $n^{th}$ row of $\mathbf{R}_{ij}$. 
    
\textit{Hidden-State Distillation.} Similarly, we can generate relation matrices for hidden states, \emph{i.e.}, the features output by MLP. Denoting the hidden states of a transformer layer by $\mathbf{H}\in\mathbb{R}^{N\times d}$, the hidden-state distillation loss based on relation matrices is defined as
    \begin{equation}
        \mathcal{L}_{hddn}=\frac{1}{N}\sum_{n=1}^NCE(\mathbf{R}_{H,n}^s, \mathbf{R}_{H,n}^t),
    \end{equation}
    where $\mathbf{R}_{H,n}$ indicates the $n^{th}$ row of $\mathbf{R}_H$, which is  computed by $\mathbf{R}_H=\text{softmax}(\mathbf{H}\mathbf{H}^T/\sqrt{d})$.

    Based on our observation that only using prediction soft labels can yield better performance than using both the prediction and ground truth labels 
    (see the ablation in Sec.~\textcolor{red}{\ref{sec: analysis}}), thus the final distillation objective function is formulated as
    \vspace{-4mm}
    \begin{align}
        \mathcal{L}_{train}=\mathcal{L}_{pred}+\beta\mathcal{L}_{attn} 
        + \gamma\mathcal{L}_{hddn}, \label{eq: loss_distill}
    \end{align}
    where $\beta$ and $\gamma$ are hyperparameters with default values 
    of 1 and 0.1 respectively, unless otherwise specified.
    
\subsection{Compression Pipeline}
Our compression pipeline includes two phases:

\textit{Phase 1: Generating compact architectures with weight transformation.} Given a large pre-trained vision transformer model, we first share the parameters across every-$K$ adjacent transformer layers except the LayerNorm \cite{LN}. Then we apply weight transformation to each layer by inserting a tiny linear layer before and after the \emph{softmax} layer, as defined in Eq.~(\ref{eq:t_mha_before}-\ref{eq:t_mha_after}). Furthermore, we introduce a depth-wise convolutional layer for MLP. These linear layers and transformation blocks are not shared.

\textit{Phase 2: Training the compressed models with weight distillation.} In this step, we apply the proposed weight distillation method to transfer knowledge from the large pre-trained models to small ones, using the objective function defined in Eq.~(\ref{eq: loss_distill}).~Such distillation inside the transformer modules allows the student network to reproduce the behaviors of the teacher network\cite{tinybert}, thus extracting more useful knowledge from the large-scale pre-trained models. Note that it is only performed when both the teacher and student models are transformer architectures. In other cases where the architectures of the student and teacher are heterogeneous, we only preserve the prediction-logit distillation. 

     \begin{figure}[t]
     \vspace{-3mm}
        \centering
        \subfloat[][DeiT-S]{
        \includegraphics[height=2.0cm]{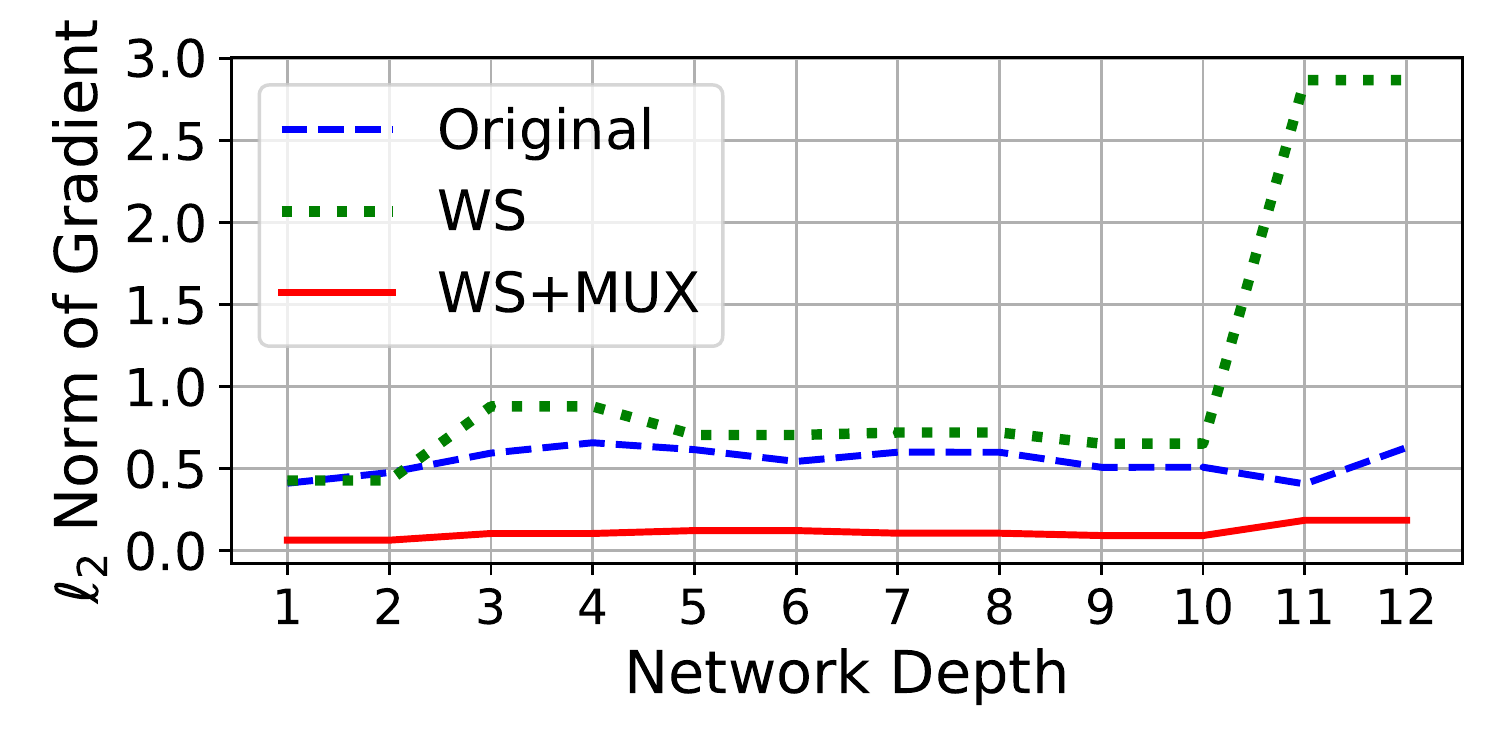}
        }
        \hspace{-5mm}
        \centering
        \subfloat[][Swin-B]{
        \vspace{-2.1mm}
        \includegraphics[height=2.2cm]{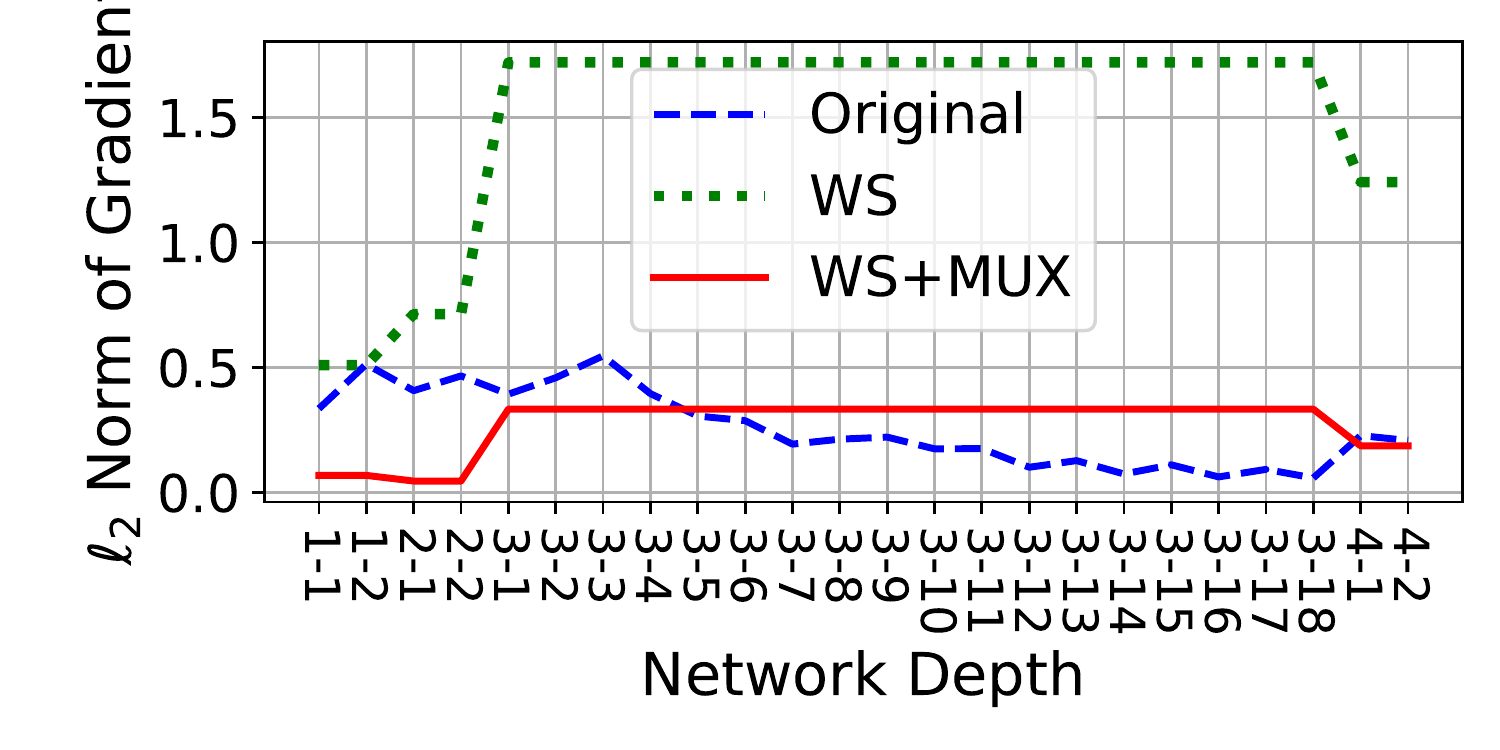}
        }
    \vspace{-4mm}
        \caption {
        Comparisons of $\ell_2$-norm of gradients during training among the models with weight sharing (WS), and with both weight sharing and multiplexing (WS+MUX), and the original one. 
        }
        \vspace{-2mm}
    
        \label {fig:grad_norm}
    \end{figure}
    
    \begin{figure}[t]
        \centering
        \subfloat[][DeiT-S]{
        \includegraphics[height=2.0cm]{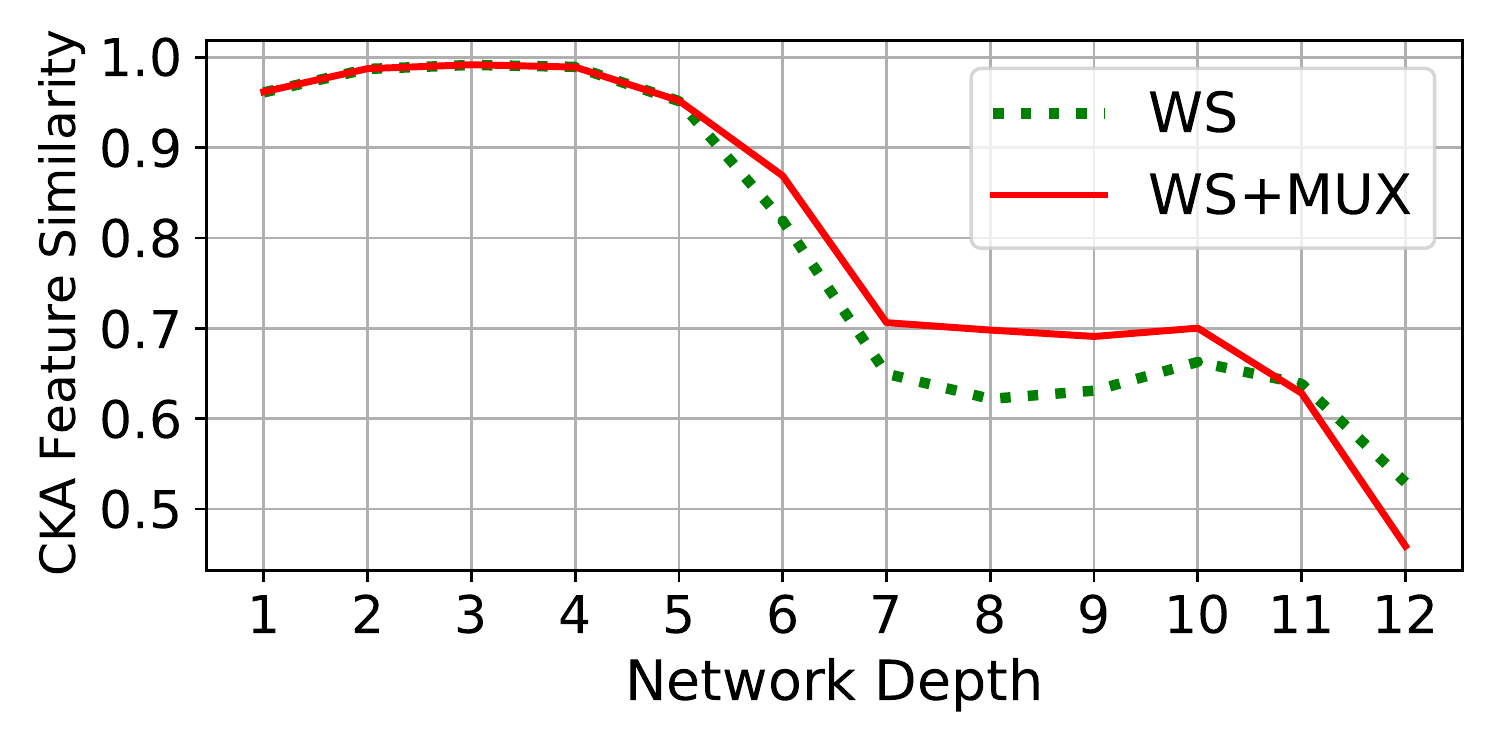}
        }
        \hspace{-3mm}
        \centering
        \subfloat[][Swin-B]{
        \includegraphics[height=2.0cm]{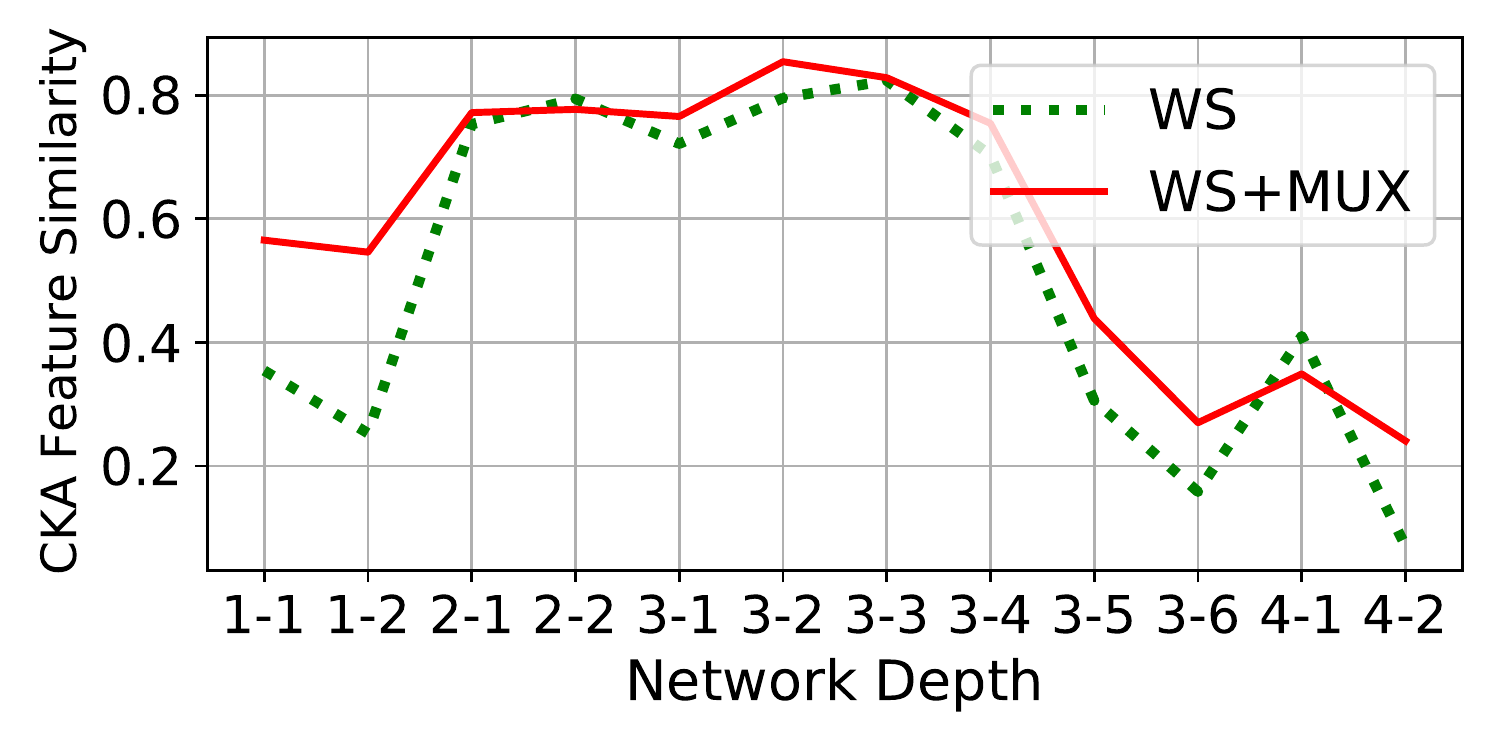}
        }
     \vspace{-4mm}
        \caption {
        Comparisons of feature similarity with respect to the original model by CKA~\cite{cka} between the models with weight sharing (WS), and with weight sharing and multiplexing (WS+MUX).
        }
        \vspace{-5mm}
    
        \label {fig:feat_cka}
    \end{figure}
    
    \section{Experiments}
    In this section, we first provide an analysis of weight sharing in vision transformers, followed by experiments on the effects of the proposed weight transformation and weight distillation methods. Next, we show the parameter efficiency of our method by comparing with other state-of-the-art models. Finally, we demonstrate the transferability of the compression models on downstream tasks.
    
    \vspace{-1mm}
    \subsection{Implementation Details}

\textit{Architectures for Compression.} 
In the case of hierarchical models such as Swin~\cite{Swin}, we only share parameters of transformer layers in each stage, due to the unaligned parameter dimension in different stages (caused by feature down-sampling), and generate a series of compact models named \textit{Mini-Swins}. For DeiT, one of the popular isomorphic models, we consider it as a single-stage vision transformer and create \textit{Mini-DeiTs}. We make several modifications on DeiT: First, we remove the [class] token. The model is attached with a global average pooling layer and a fully-connected layer for image classification. We also utilize relative position encoding to introduce inductive bias to boost the model convergence~\cite{deit, irpe}. Finally, based on our observation that transformation for FFN only brings limited performance gains in DeiT, we remove the block to speed up both training and inference.~The results of MiniViT based on the original DeiT~\cite{deit} are presented in the \textit{supplementary materials} for comprehensive comparisons.

\textit{Training Settings.} We train our models from scratch on ImageNet-1K~\cite{imagenet} by directly inheriting the hyper-parameters from DeiT~\cite{deit} or Swin transformers~\cite{Swin} except for the drop path rate, which are set to be 0.0/0.0/0.1 for DeiT-Ti/S/B and 0.0/0.1/0.2 for Swin-T/S/B, respectively.~The data augmentation techniques include RandAugment~\cite{randaug}, Cutmix~\cite{cutmix}, Mixup~\cite{mixup}, and random erasing. RepeatAug~\cite{repeat_aug} is only applied in DeiT~\cite{deit}. 
In the downstream tasks, we fine-tune the models for 30 epochs at $384^2$ resolution. Specifically, the position encoding is resized with bicubic interpolation. The AdamW~\cite{adamw} optimizer is applied with weight decay $10^{-8}$ and a cosine scheduler, batch size 256. The learning rates are $2.5\times10^{-6}$ and $10^{-5}$ for DeiT and Swin, respectively.
All models are implemented using PyTorch~\cite{pytorch} and Timm library~\cite{timm}, and trained for 300 epochs with 8 NVIDIA Tesla V100 GPUs. 

For compressing Swin~\cite{Swin}, we adopt the ImageNet-22k~\cite{imagenet} pre-trained Swin-B with 88M parameters as the teacher model. Thus, our compressed models can still learn the knowledge of the large-scale ImageNet-22k data by distillation without requiring the access to the dataset. Similarly,
for DeiT \cite{deit}, 
we use RegNet-16GF~\cite{radosavovic2020designing} with 84M parameters as the teacher, 
and only perform prediction-logit distillation due to heterogenous architectures between CNN and ViT.
We share the consecutive two layers except Swin-T where all layers in each stage are shared. For the depth-wise convolution in weight transformation, we set the kernel size to be the same as the window size in Swin~\cite{Swin} and the stride as 1. Besides, same-padding is used. 
     
    \begin{table}[t]
    \vspace{-0.6cm}
      \centering
      \setlength{\tabcolsep}{8pt}{
      \scalebox{0.75}{
      
      \begin{tabular}{l|c|ccc|ccc}
         \Xhline{2\arrayrulewidth}
        \multirow{2}*{Model} & \multirow{2}*{\#} & \multirow{2}*{WS} & \multicolumn{2}{c|}{MUX} & \multirow{2}*{\#Params} & Top-1 & Top-5 \\
        ~ & ~ & ~ & WD & WT & ~ & Acc(\%) & Acc(\%) \\ 
         \Xhline{2\arrayrulewidth}
        \multirow{5}{*}{\swintiny} & 1 &  & & & 28M & 81.2 & 95.5\\
        ~ & 2 & \cmark &&&12M& 79.0&94.4\\
        ~ & 3 & \cmark &\cmark &&12M& 79.8 &95.1 \\
         ~ & 4 & \cmark & &\cmark&12M& 79.2 &94.3\\
         ~ &  5 & \cmark &\cmark &\cmark&12M& 81.4&95.8\\
         \hline
         \multirow{5}{*}{\deitsmall} & 6 &  & & & 22M& 79.9 & 95.0 \\
         ~ & 7 & \cmark & &&11M& 78.3 & 94.4\\
         ~ & 8 & \cmark & \cmark & & 11M & 80.1 & 95.1 \\ 
         ~ &  9 & \cmark & &\cmark&11M& 79.3&94.8\\
         ~ &  10 & \cmark &\cmark &\cmark&11M& 80.7 & 95.4\\
        \Xhline{2\arrayrulewidth}
      \end{tabular}
      }}
      \vspace{-3mm}
      \caption{Component-wise analysis on ImageNet-1K~\cite{imagenet}. WS: Weight Sharing, WD: Weight Distillation, WT: Weight Transformation, MUX: Weight Multiplexing including both WD and WT.}
      \label{tab:component}
      \vspace{-1mm}
    \end{table}
    
    \begin{table}[t]
    \vspace{-0.3cm}
      \centering
      \setlength{\tabcolsep}{7pt}{
      \scalebox{0.75}{
      \begin{tabular}{c|lcc|lcc}
         \Xhline{2\arrayrulewidth}
       Sharing & \multicolumn{3}{c|}{\swintiny} & \multicolumn{3}{c}{\deitbase} \\
       Strategy & \#Params & Top-1 & Top-5 & \#Params & Top-1 & Top-5 \\
         \Xhline{2\arrayrulewidth}
         original & 28M &81.2&95.5& 86M&81.8&95.6\\
        every-2 &16M{\textcolor{YellowGreen}{\scriptsize{($1.8\times$)}}}&82.2& 96.2& 44M{\textcolor{YellowGreen}{\scriptsize{($2.0\times$)}}} & 83.2 & 96.5 \\
        all layers &12M{\textcolor{YellowGreen}{\scriptsize{($2.3\times$)}}}&81.4&95.8 & \ \ 9M{\textcolor{YellowGreen}{\scriptsize{($9.7\times$)}}} & 79.8 & 94.9 \\
        \Xhline{2\arrayrulewidth}
      \end{tabular}
    }}
    \vspace{-3mm}
      \caption{Ablation study on the number of sharing blocks on Imagenet-1K~\cite{imagenet}. Fisrt row: the original model. Second row: sharing two consecutive layers. Third-row: sharing all layers of DeiT~\cite{deit} and in each stage of Swin~\cite{Swin}.}
      \label{tab:sharing_times}
      \vspace{-5mm}
    \end{table}

    \subsection{Analysis and Ablation}
    \label{sec: analysis}
    \textit{Analysis of Weight Sharing (WS)}.
    When directly applying weight sharing as in Eq.~(\ref{eq: weight_sharing}), the training process collapses, and the model suffers from severe performance drop.
    
    First, we investigate the training stability. We share every two layers in DeiT-S while sharing all layers in each stage of Swin-B.
    As shown in Fig.~\ref{fig:grad_norm}, the $\ell_2$-norm of gradients in DeiT-S and Swin-B after weight sharing becomes much larger, indicating a fast change of magnitude of weights. Furthermore, weight sharing causes fluctuations in the gradient norm across different layers. This may lead to different optimization paces of layers. In particular, some layers are updated quickly, whereas the other parts are hardly optimized, making the model likely to converge to a bad local optimum or even diverge in training. Therefore, strictly identical weights shared across layers lead to training instability. However, our weight multiplexing method can both reduce the gradient norm and increase the smoothness across layers by introducing transformations to improve parameter diversity, promoting a more stable training process.
    
    \begin{table}[t]
    \vspace{-0.6cm}
      \centering
      \setlength{\tabcolsep}{4pt}{
      \scalebox{0.75}{
      \begin{tabular}{l|cccc|cc}
         \Xhline{2\arrayrulewidth}
         Model& GT & $\mathcal{L}_{pred}$ &$\mathcal{L}_{attn}$  & $\mathcal{L}_{hddn}$ & Top-1 & Top-5 \\ 
         \Xhline{2\arrayrulewidth}
         \swinbase (22k) (Teacher) &  \cmark& &&& 85.2 & 97.5\\
         \hline
         {Mini-\swintiny} w/o distillation & \cmark & &&& 79.2& 94.3\\
         \hline
        \multirow{6}{*}{Mini-\swintiny} &  &\cmark &&& 81.2 &95.6\\
        ~ &  \cmark &\cmark &&&80.9 &95.5\\
          &  \ &\cmark &\cmark&&81.4&95.7\\
         ~ &    &\cmark &&\cmark& 81.5&95.8\\
         ~ &  &\cmark &\cmark&\cmark& 81.4&95.8\\
        \Xhline{2\arrayrulewidth}
      \end{tabular}
      }
      }
      \vspace{-3mm}
      \caption{Ablation study on different distillation losses on ImageNet-1K~\cite{imagenet}. We use Swin-B~\cite{Swin} pre-trained on ImageNet-22K~\cite{imagenet} as the teacher model. The weights for all losses are 1 except using both GT and $\mathcal{L}_{pred}$, where both weights are 0.5.
      }
      \label{tab:dis_loss}
      \vspace{-6mm}
    \end{table}
    
    As for performance analysis, we present a comparison of feature similarities with CKA between weight sharing and weight multiplexing in Fig.~\ref{fig:feat_cka}. Higher CKA values indicate more similar feature representations between two models \cite{cka}, thus achieving similar performance. We observe that both DeiT and Swin suffer from large deviations of feature representations after applying weight sharing, especially in the last several layers, which may be one of the reasons for performance drops brought about by weight sharing. Nevertheless, our proposed weight multiplexing method can improve the similarities. 
    
    \begin{table}[t]
    
    \vspace{-0.6cm}
        \begin{threeparttable}
        \setlength{\tabcolsep}{2pt}{
    \scalebox{0.75}{
    \begin{tabular}{c|c|c|ccc|c}
        \Xhline{2\arrayrulewidth}
        \multirow{2}*{Model} & \#Param. & MACs & IN-1k & IN-Real & IN-V2 & \multirow{2}*{Input} \\
        ~ & (M) & (B) & Acc (\%) & Acc (\%) & Acc (\%)  & ~ \\
        \Xhline{2\arrayrulewidth}
        \multicolumn{6}{c}{Convnets} \\
        \hline
        ResNet-50~\cite{resnet, timm} & 25 & 4.1
        & 69.8 & 77.3 & 57.1 & $224^2$ \\
        RegNetY-16GF~\cite{radosavovic2020designing} & 84 & 15.9 & 82.9 & 88.1 & 72.4 & $224^2$ \\
    
        EfficientNet-B1~\cite{efficientnet} & 8 & 0.7 & 79.1 & 84.9 & 66.9 & $240^2$ \\
        EfficientNet-B5~\cite{efficientnet} & 30 &  9.9 & 83.6 & 88.3 & 73.6 & $456^2$ \\
        \hline
        \multicolumn{7}{c}{Transformers} \\
        \hline
        ViT-B/16~\cite{ViT} & 86 & 55.6 & 77.9 & 83.6 & - & $384^2$ \\
        ViT-L/16~\cite{ViT} & 307 & 191.5 & 76.5 & 82.2 & - & $384^2$ \\
         \hline
         
          VTP (20\%)~\cite{VTP} & 67 & 13.8 & 81.3 & - & - & $224^2$ \\
       VTP (40\%)~\cite{VTP} & 48 &10.0 & 80.7 & - & - & $224^2$ \\
        
        \hline

         AutoFormer-T~\cite{autoformer} & 6 & 1.3 & 74.7 & - & - & $224^2$ \\
        AutoFormer-S~\cite{autoformer} & 23 & 5.1 & 81.7 & - & - & $224^2$ \\
        AutoFormer-B~\cite{autoformer} & 54 & 11.0 & 82.4 & - & - & $224^2$ \\
        \hline
         S$^2$ViTE-T~\cite{S2VIT} & 4 & 1.0 & 70.1 & - & - & $224^2$ \\
       S$^2$ViTE-S~\cite{S2VIT} & 15 &3.1 & 79.2 & - & - & $224^2$ \\
        S$^2$ViTE-B~\cite{S2VIT} & 57 & 11.8 & 82.2 & - & - & $224^2$ \\
        \hline

        DeiT-Ti~\cite{deit} & 5 & 1.3 & 72.2 & 80.1 & 60.4 & $224^2$ \\
        DeiT-S~\cite{deit} & 22 & 4.6 & 79.9 & 85.7 & 68.5 & $224^2$ \\
        DeiT-B~\cite{deit} & 86 & 17.6 & 81.8 & 86.7 & 71.5 & $224^2$ \\
        DeiT-B$\uparrow$384~\cite{deit} & 87 & 55.6 & 82.9 & 87.7 & 72.4 & $384^2$ \\
        \deitbasedisup384~\cite{deit} & 88 & 55.7 & 84.5 & 89.0 & 74.8 & $384^2$ \\
        \hline

        Mini-DeiT-Ti~\textbf{(ours)} & 3{\textcolor{YellowGreen}{\scriptsize{($1.7\times$)}}}  & 1.3 & 72.8{\textcolor{YellowGreen}{\scriptsize{(+0.6)}}} &83.5{\textcolor{YellowGreen}{\scriptsize{(+3.4)}}} & 61.3{\textcolor{YellowGreen}{\scriptsize{(+0.9)}}} & $224^{2}$ \\
        
        Mini-DeiT-S~\textbf{(ours)} & 11{\textcolor{YellowGreen}{\scriptsize{($2.0\times$)}}}  & 4.7 & 80.7{\textcolor{YellowGreen}{\scriptsize{(+0.8)}}} & 88.4{\textcolor{YellowGreen}{\scriptsize{(+2.7)}}} & 69.5{\textcolor{YellowGreen}{\scriptsize{(+1.0)}}} & $224^{2}$ \\
        
        Mini-DeiT-B~\textbf{(ours)} & 44{\textcolor{YellowGreen}{\scriptsize{($2.0\times$)}}}  & 17.7 & 83.2{\textcolor{YellowGreen}{\scriptsize{(+1.4)}}} & 89.6{\textcolor{YellowGreen}{\scriptsize{(+2.9)}}} & 73.0{\textcolor{YellowGreen}{\scriptsize{(+1.5)}}} & $224^{2}$ \\
        
        Mini-DeiT-B$\uparrow$384~\textbf{(ours)} & 44{\textcolor{YellowGreen}{\scriptsize{($2.0\times$)}}} & 56.9 & 84.7{\textcolor{YellowGreen}{\scriptsize{(+1.8)}}} & 89.9{\textcolor{YellowGreen}{\scriptsize{(+2.2)}}} & 75.2{\textcolor{YellowGreen}{\scriptsize{(+2.8)}}} & $384^{2}$ \\
        \hline
        Swin-B~(22k)~\cite{Swin} & 88 & 15.4 & 85.2 & 89.2 & 75.3 & $224^2$ \\
        Swin-B$\uparrow$384~(22k)~\cite{Swin} & 88 & 47.1 & 86.4 & 90.0 & 76.6 & $384^2$ \\
        \hline
        Swin-T~\cite{Swin} & 28 & 4.5 & 81.2 & 86.6 & 69.6 & $224^2$ \\
        Swin-S~\cite{Swin} & 50 & 8.7 & 83.2 & 87.6 & 71.9 & $224^2$ \\
        Swin-B~\cite{Swin} & 88 & 15.4 & 83.5 & 87.8 & 72.5 & $224^2$ \\
        Swin-B$\uparrow$384~\cite{Swin} & 88 & 47.1 & 84.5 & 88.6 & 73.2 & $384^2$ \\
        \hline
        Mini-Swin-T~\textbf{(ours)} & 12{\textcolor{YellowGreen}{\scriptsize{($2.3\times$)}}} & 4.6 & 81.4{\textcolor{YellowGreen}{\scriptsize{(+0.2)}}} & 87.1{\textcolor{YellowGreen}{\scriptsize{(+0.5)}}} & 70.5{\textcolor{YellowGreen}{\scriptsize{(+0.9)}}} & $224^{2}$ \\
        Mini-Swin-S~\textbf{(ours)} & 26{\textcolor{YellowGreen}{\scriptsize{($1.9\times$)}}} & 8.9 & 83.6{\textcolor{YellowGreen}{\scriptsize{(+0.4)}}} & 88.7{\textcolor{YellowGreen}{\scriptsize{(+1.1)}}} & 73.8{\textcolor{YellowGreen}{\scriptsize{(+1.9)}}} & $224^{2}$ \\
        Mini-Swin-B~\textbf{(ours)} & 46{\textcolor{YellowGreen}{\scriptsize{($1.9\times$)}}} & 15.7 & 84.3{\textcolor{YellowGreen}{\scriptsize{(+0.8)}}} & 89.0{\textcolor{YellowGreen}{\scriptsize{(+1.2)}}} & 74.4{\textcolor{YellowGreen}{\scriptsize{(+1.9)}}} & $224^{2}$\\
        Mini-Swin-B$\uparrow$384~\textbf{(ours)} & 47{\textcolor{YellowGreen}{\scriptsize{($1.9\times$)}}} & 49.4 & 85.5{\textcolor{YellowGreen}{\scriptsize{(+1.0)}}} & 89.5{\textcolor{YellowGreen}{\scriptsize{(+0.9)}}} & 76.1{\textcolor{YellowGreen}{\scriptsize{(+2.9)}}} & $384^2$ \\
        
        \Xhline{2\arrayrulewidth}
        \end{tabular}
    }}
        \end{threeparttable}
        \vspace{-0.3cm}
        \caption {
        MiniViT Top-1 accuracy on ImageNet-1K~\cite{imagenet}, Real~\cite{imagenet_real} and V2~\cite{imagenet_v2} with comparisons to state-of-the-art models. 
        Our MiniViTs consistently outperform existing transformer-based visual models and CNNs with fewer parameters. $\uparrow$ denotes fine-tuning with $384^2$ resolution.
        }
        \label {tab:sota_cls}
    \vspace{-6mm}
    \end{table}
    
    \textit{Component-wise Analysis. } 
    We evaluate the effects of different components in our proposed weight multiplexing method on ImageNet-1K~\cite{imagenet}, and report the results in Tab.~\ref{tab:component}. Our baselines are the official Swin-T~\cite{Swin} and DeiT-S~\cite{deit} models. After applying the weight-sharing method, the numbers of parameters of both models are halved, whereas the accuracy also decreases by 2\% (\#1 \emph{vs.} \#2, \#6 \emph{vs.} \#7). The performance can be improved by either applying weight distillation (\#2 \emph{vs.} \#3, \#7 \emph{vs.} \#8) or weight transformation (\#2 \emph{vs.} \#4, \#7 \emph{vs.} \#9), which demonstrates their individual effectiveness. It is notable that weight transformation only introduces a few parameters. Furthermore, when combining all three components, MiniViTs can achieve the best accuracy (\#5 and \#10), which even outperforms the original Swin-T and DeiT-S models.

    \textit{Number of Sharing Blocks.} We study the performance of MiniViTs in different sharing settings. In Tab.~\ref{tab:sharing_times}, 
    sharing every two blocks in each stage of Swin-T or DeiT-B can significantly reduce the number of parameters from 28M to 16M, and 86M to 44M, respectively, whereas the Top-1 accuracy becomes 1\% higher. Note that we consider DeiT as a single-stage vision transformer. In the extreme case where all blocks in each stage are shared, Mini-Swin-T can still outperform the original model with only 43\% of the parameters. Mini-DeiT-B can achieve 90\% parameter reduction with only a 2\% performance drop. The results indicate that parameter redundancy exists in vision transformers and our proposed MiniViT is effective in improving parameter efficiency. Moreover, MiniViT is configurable to satisfy various requirements on model size and performance.
  
    \textit{Distillation Losses.} We investigate the effectiveness of different types of distillation in MiniViT. 
    When using both ground truth (GT) and prediction soft labels as loss,
    we set the trade-off weights to 0.5. We use Swin-T as our baseline. As shown in Tab.~\ref{tab:dis_loss},
    compared to only using the prediction loss,
    the extra GT labels leads to a 0.3\% performance drop for Swin, which is due to the degradation of learning ability brought by weight sharing.
    Furthermore, we also observe around a 0.2\% 
    improvement in accuracy after applying the attention-level and hidden-state distillation, indicating the effectiveness of our proposed distillation method. 

    \begin{table}[t]
    \vspace{-0.6cm}
      \centering
      \setlength{\tabcolsep}{6pt}{
      \scalebox{0.72}{
      \begin{tabular}{l|c|c|ccccc}
         \Xhline{2\arrayrulewidth}
        Model& \#Params & \rotatebox[origin=l]{90}{ImageNet1k} & \rotatebox[origin=l]{90}{CIFAR-10} & \rotatebox[origin=l]{90}{CIFAR-100} &\rotatebox[origin=l]{90}{Flowers} & \rotatebox[origin=l]{90}{Cars} & \rotatebox[origin=l]{90}{Pets} \\ 
         \Xhline{2\arrayrulewidth}
         Grafit ResNet-50~\cite{grafit} & 25M & 79.6 & - & - & 98.2 & 92.5 &-\\
       EfficientNet-B5~\cite{efficientnet} & 30M & 83.6 & 98.7 & 91.1 & 98.5 &-&-\\
       EfficientNet-B7~\cite{efficientnet} & 66M & 84.3 & 98.9 & 91.7 & 98.8 & 94.7 & - \\
        \hline
        ViT-B/32~\cite{ViT}&86M & 73.4&97.8&86.3&85.4&-&92.0\\
        ViT-B/16~\cite{ViT}&86M & 77.9&98.1&87.1&89.5&-&93.8\\
        NViT-T~\cite{nvit}&6.4M &73.9&98.2&85.7&-&-&-\\
        NVP-T~\cite{nvit} &6.9M&76.2 & 98.3&85.9&-&-&-\\
    \deitbase~\cite{deit} & 86M & 81.8 & 99.1 &  90.8  & 98.4 &  92.1    & - \\
    \deitbaseup384~\cite{deit}    & 87M & 83.1 & 99.1 &  90.8  & 98.5 &  93.3    &  - \\
    \deitbasedis~\cite{deit}   & 87M & 83.4 & 99.1  &  91.3  & 98.8 & 92.9 & - \\
    \deitbasedisup384~\cite{deit} & 88M & 84.4 & 99.2 & 91.4   & 98.9 & 93.9 & -\\
        \hline
        
        Mini-DeiT-B$\uparrow$384 & 44M & 84.7 & 99.3 & 91.5 & 98.3 & 93.8 & 95.5 \\
        
        \Xhline{2\arrayrulewidth}
      \end{tabular}
      }}
      \vspace{-0.3cm}
      \caption{MiniViT results on downstream classification datasets.
      }
      \label{tab:downstream_task}
      \vspace{-6mm}
    \end{table}

   \begin{table*}[t]
    \vspace{-0.6cm}
    \centering
      \setlength{\tabcolsep}{11pt}{
      \scalebox{0.75}{
      
    \begin{tabular}{c|cc|ccc|c|cccccc}

        \Xhline{2\arrayrulewidth}
        \# & Backbone & \#Params & WT & WD & Det KD & ImageNet Top1-Acc & $AP$ & $AP_{50}$ & $AP_{75}$ & $AP_S$ & $AP_M$ & $AP_L$ \\
        \Xhline{2\arrayrulewidth}
        1  & Swin-T~\cite{Swin} & 28M &  &  &  & 81.2 &  48.1 & 67.1 & 52.1 & 31.1 & 51.2 & 63.5 \\
        \hline
        2 & Mini-Swin-T & 12M &  &  &  & 79.0 & 46.5 & 65.5 & 50.6 & 29.9 & 50.0 & 61.6 \\
        3 & Mini-Swin-T & 12M & \cmark &  &  & 79.2 & 47.5 & 66.1 & 51.8 & 30.8 & 50.6 & 62.2 \\
        4 & Mini-Swin-T & 12M & \cmark & \cmark &  & 81.4 & 47.8 & 66.5 & 51.9 & 30.4 & 51.3 & 63.2 \\
        5 & Mini-Swin-T & 12M & \cmark & \cmark & \cmark & 81.4 & 48.6 & 67.2 & 52.6 & 31.0 & 52.4 & 64.2 \\ 
        \Xhline{2\arrayrulewidth}
    \end{tabular}}}
            \vspace{-0.3cm}
        \caption {Comparison on COCO~\cite{coco} object detection using Cascade Mask R-CNN~\cite{cascade_rcnn, mask_rcnn}. We replace the original backbone with our compressed models, and report the number of parameters of the backbone. We train detectors for 12 epochs.
        }
    
        \label {tab:det_comp}
        \vspace{-0.6cm}
    \end{table*}
    
    \vspace{-0.1cm}
    \subsection{Results on ImageNet}
    \vspace{-0.1cm}

    We compare our proposed MiniViT models for diverse parameter sizes with the state-of-the-art ones on ImageNet-1K~\cite{imagenet}, Real~\cite{imagenet_real} and V2~\cite{imagenet_v2}. The top-1 accuracy is reported in Tab.~\ref{tab:sota_cls}. Note that our MiniViTs are trained from scratch on ImageNet-1K only without using the large-scale ImageNet-22K dataset.
    It is clear that the Mini-Swin and Mini-DeiT model families, compressed over Swin transformers \cite{Swin} and DeiT \cite{deit}, respectively, achieve accuracy improvements with only half as many parameters. In particular, using ~46M parameters, our Mini-Swin-B performs 0.8\% accuracy higher than Swin B on ImageNet-1K. Besides, our Mini-DeiT-B reduces 50\% parameter amount and achieves a 1.8\% performance improvement. Moreover, on ImageNet-Real, our Mini-DeiTs achieve 2\% better performance than DeiTs, while Mini-Swins are also superior to Swins by around 1\%. On ImageNet-V2, MiniViTs can outperform the original models up to 3\% (Base$\uparrow$384).
    
    Compared to other efficient vision transformer methods, MiniViT is also competitive. Specifically, Mini-DeiT-B, with only 44M parameters, achieves 1.0\% and 2.5\% higher Top-1 accuracy than S$^2$ViTE-B (57M) \cite{S2VIT},  and VTP (40\%) (48M) \cite{VTP}, respectively. 
    
    Our Mini-DeiTs can also outperform automatical nearul architecture search methods. In particular, Mini-DeiT-B uses only 44M parameters to achieve 0.8\% better accuracy than AutoFormer-B (54M) \cite{autoformer}. Our tiny model, Mini-DeiT-Ti, still has comparable performance to AutoFormer-T, while the model size is merely 3M.

    \subsection{Transfer Learning Results}
    
    \textit{Image Classification. } 
    We transfer MiniViT to a collection of commonly used recognition datasets: (1) CIFAR-10 and CIFAR-100~\cite{cifar}; (2) fine-grained classification: Flowers~\cite{flowers}, Stanford Cars~\cite{stanford_cars}, and Oxford-IIIT Pets~\cite{pet}. Following the fine-tuning settings of DeiT~\cite{deit}, an SGD optimizer is adopted with learning rate $5\times10^{-3}$, batch size 256, weight decay $10^{-4}$, and disabled random erase, except Cars~\cite{stanford_cars}, using AdamW~\cite{adamw} with learning rate $10^{-3}$, weight decay $5\times10^{-2}$ and random erase. We train models for 300 epochs on Flowers and Cars, and 1000 on others. Tab.~\ref{tab:downstream_task}
 shows the results of Top-1 accuracy. Compared to the state-of-the-art ConvNets and transformer-based models, Mini-DeiT-B$\uparrow$384 achieves comparable or even better results on all datasets, only using 44M parameters. 
    
    \textit{Object Detection.} We also investigate the transferability of MiniViT to the COCO 2017 detection dataset~\cite{coco}. We use Cascade R-CNN~\cite{cascade_rcnn} with Swin-T~\cite{Swin} as our baseline. 
   For distillation during fine-tuning (Det KD), Cascade R-CNN with Swin-B is adopted as the teacher model, and a mean square error (MSE) loss on the backbone outputs of the student and teacher models is utilized with weight 0.1. We follow the same training techniques with Swin-based Cascade R-CNN~\cite{Swin}. As shown in Tab.~\ref{tab:det_comp}, (1) weight-sharing can cause a 1.6 AP decrease, although it reduces the number of parameter; (2) models with weight transformation can transfer well, with only a 0.6 AP decrease but fewer parameters; (3) MiniViT, combining weight sharing, transformation, and distillation, can achieve comparable AP performance to the baseline with 57\% parameter reduction on the backbone, and can achieve the best result after coupled with detection distillation in the fine-tuning stage.

\vspace{-0.2cm}    
\section{Related Work}

\textit{Vision Transformer.} Transformers were originally proposed for language modeling~\cite{vaswani2017attention}, and recently applied in computer vision. It has shown promising potential on a variety of tasks, such as recognition, detection, and segmentation~\cite{DETR, ViT, liang2020polytransform}.
Dosovitskiy et al. first introduced the ViT model \cite{ViT}, a pure transformer architecture for visual
recognition pre-trained on large-scale data.
This work inspired a large amount of follow-up approaches \cite{deit, Swin, CrossViT,T2TViT, han2021transformer}. Among them, DeiT \cite{deit} and the Swin transformer \cite{Swin} are two representative ones. DeiT \cite{deit} demonstrates that large-scale data is not necessary when training a ViT model.
Swin transformers \cite{Swin} introduce a hierarchical structure to the ViT regime, mimicking traditional convolutional networks. Equipped with shifted windows, they have shown promising results on visual recognition and downstream tasks.

ViT models are becoming increasingly heavy and expensive, so several recent works have proposed methods for compression. One primitive way is pruning, such as removing redundant tokens \cite{dynamicvit, IA-RED2}, attention heads \cite{S2VIT}, or hidden dimensions \cite{VTP, nvit}. A recent work \cite{UVTC} combines pruning, skipping, and distillation together and proposes a unified compression framework for vision transformers. However, most existing compression literature focuses on isomorphic vision transformers, whereas their efficacy on hierarchical vision transformers remains unclear. Our method is more general, aiming at the compression of both isomorphic and hierarchical vision transformers.
    
\textit{Weight Sharing.}
The history of weight sharing can be traced back to the 1990s~\cite{ws_origin, ws_lecun, ws_hinton}. Due to its good capability in improving parameter efficiency,
weight sharing has been adopted in transformers for language tasks~\cite{Albert, LT, ws_trans_ma}. Universal transformers~\cite{UniTransformer} propose a weight sharing mechanism across both positions and time steps, yielding better performance than standard transformers~\cite{vaswani2017attention} on a variety of sequence-to-sequence tasks. After observing that the output of weight-shared models converges to a fixed point, Deep Equilibrium Models~\cite{DEQ} design a Quasi-Newton method utilizing equilibrium states to fit transformers, and have been demonstrated to outperform other deep sequence models on language tasks. Different from the previous work, Albert~\cite{Albert} found a performance drop in NLP benchmarks after sharing all weights of transformers in BERT \cite{bert}. By contrast, our work investigates the efficacy of weight sharing in vision transformers, while equipping it with transformation and distillation to further enhance the method.

\textit{Knowledge Distillation.} Distillation in a teacher-student framework, is widely used to reduce model sizes. It has been extensively studied in convolutional networks~\cite{gou2021knowledge}. However, in vision transformers, it is still under-explored. A few relevant recent works include Touvron et al.~\cite{deit} who introduce a distillation token to allow the transformer to learn from a ConvNet teacher, and Jia et al.~\cite{jia2021efficient} propose to excavate knowledge from the teacher transformer via the connection between images and patches. 
Distillation inside the transformer (e.g., MSA and MLP) has been verified to be effective in transformers for NLP \cite{tinybert, mobilebert, minilm, minilmv2}. In contrast, our work provides an initial design for attention-level and hidden-state distillation, and explores their effectiveness in vision transformer compression.

\vspace{-0.2cm}
    \section{Conclusion}
    
    We have proposed a new compression framework, \emph{i.e.} MiniViT, for vision transformers. Our method combines weight sharing, transformation, and distillation to reduce the number of parameters while achieving even better performance compared to the original models.

\noindent
    \textit{Limitations.} One limitation of MiniViT is that, despite improving the parameter efficiency, the computational cost is slightly increased compared to the classical weight sharing strategy, due to the introduced weight transformation blocks. Second, we observe that MiniViT suffers from moderate performance degradation as the compression ratio increases. In future work, we are interested in further improving both the parameter and computational efficiency.

\noindent
    \textit{Broader Impacts.} Presented in the supplementary materials.

\textbf{Acknowledgement.} We would like to thank Microsoft \href{https://github.com/microsoft/nni}{NNI} and \href{https://github.com/microsoft/pai}{OpenPAI} v-team for kind helps and supports. Thanks to Prof. Po-Ling Loh for the final proofreading.

    {\small
    \bibliographystyle{ieee_fullname}
    \bibliography{egbib}
    }
    \end{document}